\documentclass[10pt,twocolumn,letterpaper]{article}

\usepackage[pagenumbers]{cvpr} 

\usepackage{graphicx}
\usepackage{amsmath}
\usepackage{amssymb}
\usepackage{booktabs}
\usepackage{bm}

\usepackage{color}
\usepackage{xcolor}
\definecolor{citecolor}{HTML}{0071bc}
\usepackage[pagebackref=true,breaklinks=true,colorlinks,citecolor=citecolor,bookmarks=false]{hyperref}

\usepackage[capitalize]{cleveref}
\crefname{section}{Sec.}{Secs.}
\Crefname{section}{Section}{Sections}
\Crefname{table}{Table}{Tables}
\crefname{table}{Tab.}{Tabs.}

\def\cvprPaperID{2342} 
\def\confName{CVPR}
\def\confYear{2022}

\begin{document}

\title{Look Outside the Room: Synthesizing A Consistent Long-Term 3D Scene Video from A Single Image}

\author{Xuanchi Ren\\
HKUST\\
\and
Xiaolong Wang\\
UC San Diego \\
}

\twocolumn[{%
\renewcommand\twocolumn[1][]{#1}%
\maketitle
\begin{center}
\vspace{-1.5 em}
\renewcommand\arraystretch{0.5} 
\centering
\begin{tabular}{c}
\includegraphics[width=\linewidth]{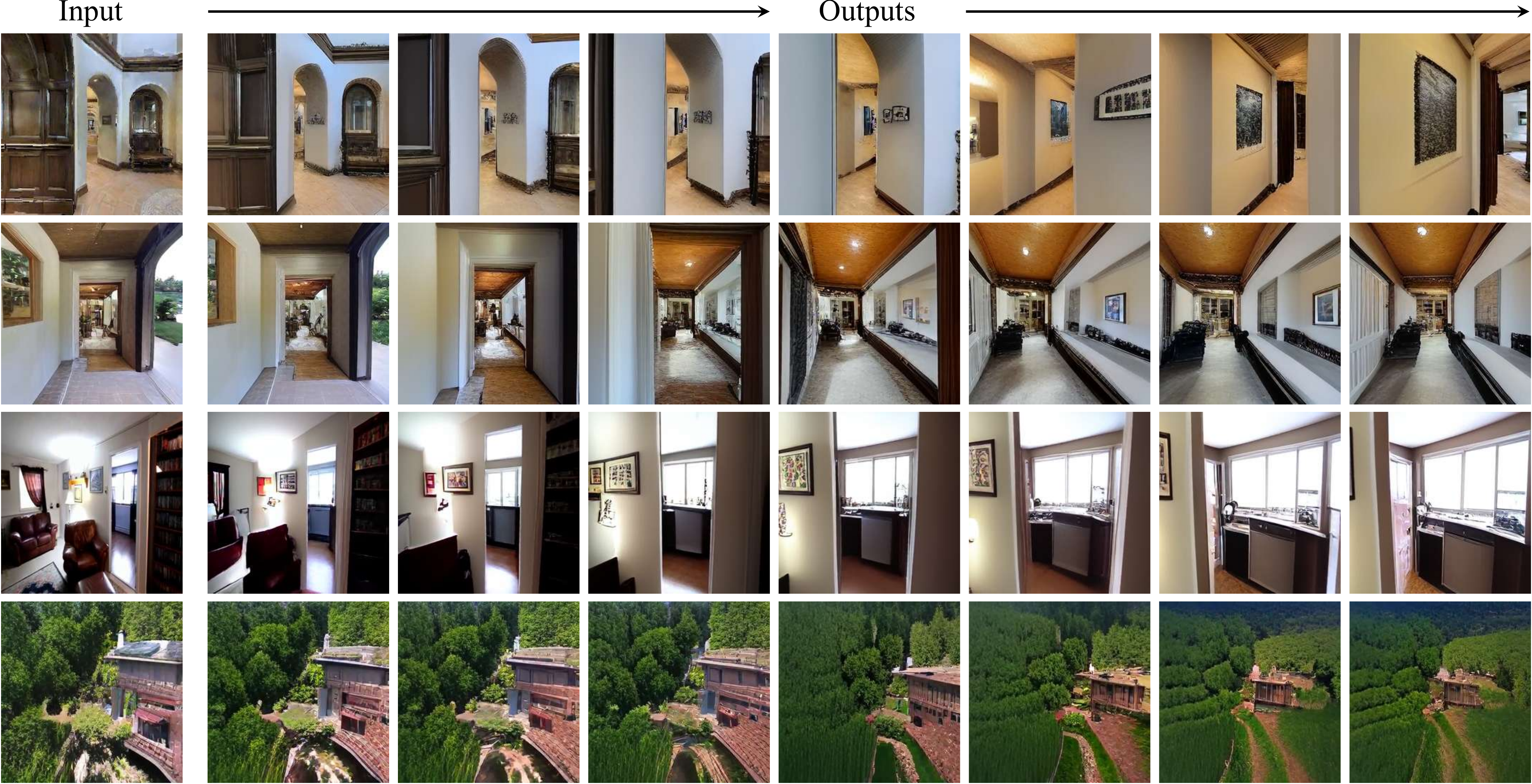} \\
\end{tabular}
\captionof{figure}{Using a single image and a camera trajectory as inputs, our method synthesizes perceptual consistent novel views, which form a long-term video. The top two rows are test images from the Matterport dataset~\cite{MP3D}, and the bottom two rows are from the RealEstate10K dataset~\cite{ZhouTFFS18}. Our model is able to synthesize with large camera changes and even walk outside a room to look into another room.}
\label{fig:teaser}
\end{center}
}]

\begin{abstract}
Novel view synthesis from a single image has recently attracted a lot of attention, and it has been primarily advanced by 3D deep learning and rendering techniques. However, most work is still limited by synthesizing new views within relatively small camera motions. In this paper, we propose a novel approach to synthesize a consistent long-term video given a single scene image and a trajectory of large camera motions. Our approach utilizes an autoregressive Transformer to perform sequential modeling of multiple frames, which reasons the relations between multiple frames and the corresponding cameras to predict the next frame. To facilitate learning and ensure consistency among generated frames, we introduce a locality constraint based on the input cameras to guide self-attention among a large number of patches across space and time. Our method outperforms state-of-the-art view synthesis approaches by a large margin, especially when synthesizing long-term future in indoor 3D scenes. Project page at \href{https://xrenaa.github.io/look-outside-room/}{https://xrenaa.github.io/look-outside-room/}.
\end{abstract}

\section{Introduction}
\label{sec:intro}
Single-image view synthesis has attracted a lot of attention in computer vision and computer graphics. It brings a photo to life by extrapolating beyond the input pixels and generating new pixels following the geometric structure of the scene. At the same time, the generated pixels need to be semantically coherent with the existing pixels. Current view synthesis methods which learn 3D geometric representation have shown encouraging results in generating high-quality novel views~\cite{synsin,MildenhallSTBRN20,TuckerS20}. However, these approaches can only generate views within a limited range of camera motion. For example, it will be very challenging for current approaches to synthesize what is outside the door of the room shown in the first row of  Figure~\ref{fig:teaser}.

When synthesizing images with large camera view changes, we would also expect the generated images to be consistent. That is, when we are synthesizing with a path walking towards the door in a room, we hope that the surroundings of the path should not change all the time and reveal a single underlying world.  
To this end, we propose to solve the problem extended based on view synthesis: Given a single image of the 3D scene and a \emph{long-term} camera trajectory as inputs, synthesize a \emph{consistent} video as the output. 
For example, given a single input image of a room (first row of Figure~\ref{fig:teaser}), we synthesize the video on walking towards the door, going through the door, and navigating into a hallway with a painting on the wall. Solving such a task not only has wide applications in content generation and editing but also helps build a differentiable simulator for model-based planning and control in robotics.

To solve this problem, we seek help from autoregressive models~\cite{van2016pixel,RazaviOV19,chen2020generative,gpt}, which have shown tremendous success in extrapolating the contents beyond the input image. 
For example, Rombach et al.~\cite{geogpt} proposes to use an autoregressive Transformer to implicitly perform \emph{large} geometric transformation for view synthesis. To handle the uncertainty with a large transformation, the model is trained under a probabilistic framework which allows for sampling different novel views with the same camera. While generating realistic novel views even given a large transformation, it also leads to inconsistent and diverse outputs along a given trajectory due to the probabilistic sampling. 

In this paper, to synthesize consistent long-term videos, we propose to leverage the autoregressive Transformer for sequential modeling in time with locality constraints.  Instead of learning the autoregressive model between only two views of the scene~\cite{geogpt}, our work leverages the continuity in videos and perform sequential modeling with multiple video frames. 
Given a sequence of input images $\{{x}_1, {x}_2, ..., {x}_{t-1}\}$ and the previous camera trajectory $\{{C}_2, {C}_3, ..., {C}_{t-1}\}$ and the camera for the future frame ${C}_{t}$, we provide a probabilistic framework to predict the future frame via sampling from $p(x_t | x_1, C_2, x_2, C_3, ..., x_{t-1}, C_{t} )$. By conditioning multiple frames during sampling, it ensures the consistency between generated views and historical views. 
When inference with our Transformer model, we can start with a single input image and gradually increase the inputs using the predicted frames and previous frames. 

However, it is very challenging to learn such a sequential model with the autoregressive Transformer, which uses self-attention to model a large number of relations between every two patches across space and time in the input video. To facilitate training, our key insight is that not every relational pair is equally important, and we can incorporate a locality constraint to guide the model to concentrate on the critical dependencies. Such locality constraints are introduced by the cameras. Intuitively, given a relative camera between two frames, we can roughly locate where the overlapping pixels are and where are the new pixels to synthesize. To incorporate this knowledge, we compute a bias using an MLP, which takes the relative camera as inputs, namely \emph{Camera-Aware Bias}. We add this bias to the affinity matrix while performing the self-attention operation. In this way, each patch will have a stronger bias on depending on or attending to relevant patches connected by the camera. Empirically, we find the Camera-Aware Bias not only makes the optimization much easier but also plays a vital role in enforcing the consistency between frames during generation.

We perform our experiments on multiple datasets, including the RealEstate10K~\cite{ZhouTFFS18} and Matterport3D~\cite{MP3D}, which mainly focus on 3D indoor scenes. Our model is able to synthesize new views with large camera motion, and generate a long-term video given a single image input as visualized in Figure~\ref{fig:teaser}. Our method not only outperforms state-of-the-art approaches on standard view synthesis metrics, but also achieves a significantly better gain when evaluating in terms of long-range future frames. We highlight our main contributions as follows:
\begin{itemize}
\vspace{-0.05in}
    \item A novel Transformer model on synthesizing a consistent long-term video given a single image and a trajectory as inputs. 
    \vspace{-0.1in}
    \item A novel locality constraint using Camera-Aware Bias, which facilitates optimization during learning and enforces the consistency between generated frames.
    \vspace{-0.1in}
    \item State-of-the-art performance in view synthesis. Our method outperforms baselines by a large margin on the long-term frames. 
\end{itemize}

\section{Related Work}
\noindent\textbf{Novel View Synthesis.} 
View synthesis has been a long-studied problem in computer vision and graphics. When synthesizing with multiple input views, 3D structural representations are often leveraged such as classical multi-view geometry~\cite{ChenW93, DebevecTM96, GortlerGSC96, LevoyH96, SeitzCDSS06, ZitnickKUWS04}, deep voxel representations~\cite{SitzmannTHNWZ19,LiuGLCT20}, and neural radiance fields~\cite{MildenhallSTBRN20,WangWGSZBMSF21}. Recently, researchers have also proposed to perform single-image view synthesis to bring a static photo to life~\cite{video_autoencoder,0006CH19, TulsianiTS18,YuYTK21,synsin,pixelsin,hu2021worldsheet,BoxPrior}. For example, Wiles et al.~\cite{synsin} propose to perform view synthesis using 3D point clouds as intermediate representations. While these approaches work well with small camera changes, they cannot outpaint pixels far from the given view. To perform view synthesis with large camera changes, Rombach et al.~\cite{geogpt} propose a Transformer based autoregressive model. While this approach can synthesize diverse and realistic results, it cannot synthesize consistent views along a trajectory. To seek a balance, Rockwell et al.~\cite{pixelsin} propose to leverage both 3D representation and the autoregressive models to achieve consistent view synthesis in indoor scenes with large camera changes. However, they are not able to generate a long-term future outside the door of the given room like our approach does. Our work is highly inspired with the idea of building ``Infinite Images'' in~\cite{KanevaSTAF10}. Instead of performing explicit matching through a large-scale dataset, we synthesize the novel scene by sampling with a Transformer model.

\noindent\textbf{Video Synthesis.} 
Learning to synthesize a video provides an important manner to capture the dynamics of the world. Researchers have studied synthesizing videos from a random noise vector~\cite{Tulyakov0YK18, SaitoMS17,VondrickPT16}, predicting the future frames based on one or multiple previous frames~\cite{DriveGAN, FinnGL16, MathieuCL15, WalkerDGH16, DentonB17, LiFYWLY18,walker2021predicting}, and translating one video from one modality to another~\cite{Wang0ZYTKC18,ChanGZE19, rendance20,Wang0TLCK19}. 
However, most video synthesis approaches do not consider the underline 3D geometry of the scene when predicting the pixels. Our work is mostly related to~\cite{Infinite_Nature}, which proposes an approach to synthesize a long-term video of outdoor nature environments given a single image and a trajectory as inputs. Different from them, we focus on 3D indoor scenes, which requires more structural reasoning when performing outpainting.

\noindent\textbf{Image Extrapolation and Outpainting.} Image outpainting~\cite{KrishnanTSMLBF19, YangDLYY19, WangTSJ19} synthesizes pixels beyond current input images in 2D. Specifically, our work is related to the autoregressive models~\cite{van2016pixel, MenickK19, ReedOKCWCBF17, SalimansK0K17, OordKEKVG16}, which perform outpainting the next pixels in a sequential manner. However, learning to predict pixels one by one introduces a large complexity in training and inference. Recently, Razavi et al.~\cite{RazaviOV19} propose a novel representation with Vector Quantized Variational AutoEncoder (VQ-VAE), which performs autoregressive modeling in latent space instead of pixel space. This largely reduces the complexity in sequential modeling, and it enables Generative Adversarial Networks~\cite{vqgan,lin2021infinitygan} for synthesizing high-resolution images with Transformers. Our work is highly inspired by these works. Besides forwarding only image tokens to Transformers, we also add cameras as tokens in sequential modeling similar to~\cite{geogpt}. 

\noindent\textbf{Transformers.} 
With the success of Transformer in language-modeling~\cite{tranformer, gpt, BERT}, it is also recently introduced into multiple recognition tasks in computer vision~\cite{ViT,slot_attention,DETR,swin,fan2021multiscale,arnab2021vivit,bertasius2021space}.Besides recognition tasks, it has also been widely used together with autoregressive models for image and video generations~\cite{ATISS,vqgan,lin2021infinitygan,geogpt}. 
However, it is still very challenging to optimize the self-attention module in Transformer when modeling a long sequence of visual tokens. In this paper, we propose to introduce a novel camera-aware bias as a locality constraint for better sequential modeling.  

\section{Method}
\begin{figure*}[t]
\centering
\includegraphics[width=\linewidth]{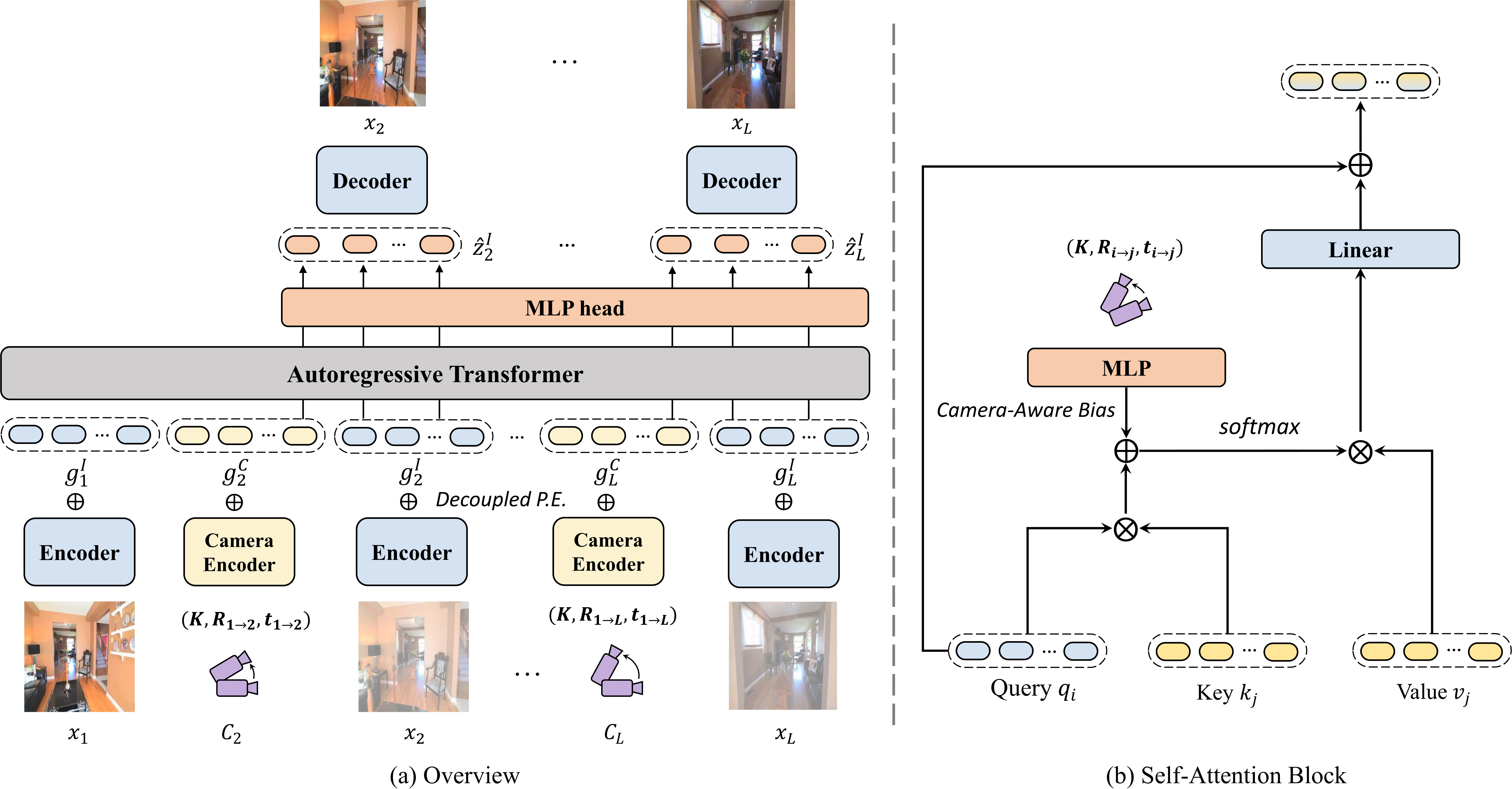}
\vspace{-1em}
\caption{
The illustration of (a) model overview, and (b) self-attention block with proposed Camera-Aware Bias.  During training, images $\{x_l \}_{l=1}^{L}$ and camera transformations $\{C_l \}_{l=2}^{L}$ are first encoded to modality-specific tokens, and a decoupled positional embedding (P.E.) is added. Tokens are then fed into an autoregressive Transformer that predicts images. 
During inference, given a single image, $x_1$ and a camera trajectory $\{C_l \}_{l=2}^{L}$, novel views can be generated autoregressively by using the Transformer.
}
\label{fig:pipeline}
\vspace{-1em}
\end{figure*}

We propose a Transformer based autoregressive model to encode and synthesize videos in a sequential manner. We will first introduce our network architecture, and our novel locality constraints using camera-aware bias for self-attention as shown in Figure~\ref{fig:pipeline}. Then we discuss the detailed training procedure. 

\subsection{Autoregressive Scene Synthesis}

Given a single input image $x_1$ together with a sequence of desirable camera transformations $\{{C}_2, {C}_3, ..., {C}_{T}\}$,  we hope to synthesize a sequence of images $\{{x}_2, ..., {x}_{T}\}$ with unconstrained length, ensuring high-quality and perceptual consistency without any 3D information.

Inspired by the success of sequential modeling in reinforcement learning~\cite{chen2021decisiontransformer}, we propose to leverage a sequential of previous frames and cameras to synthesize future novel views. To achieve this, we need to accumulate the likelihood of generating $\{x_t\}^T_{t=2}$ autoregressively as,
\begin{equation}
\begin{aligned}
 p( \{ x_{t} \}^{T}_{t=2} | x_1, \{ C_t \}_{t=2}^T ) &= \prod_{\tau , i} p( x_{\tau, i} | x_{\tau, <i}, x_{<\tau}, \{ C_t \}_{t=2}^T) \\
  &= \prod_{\tau , i} p( x_{\tau, i} | x_{\tau, <i}, x_{<\tau}, \{ C_t \}_{t=2}^\tau), \\
\end{aligned}
\label{eq:prob_1}
\end{equation}
where $\tau \in [1,T]$ indicates timestep, and $i \in [1, HW]$ indicates the index inside a flattened image coordinate. 
Based on this, we can sample $x_t$ from from the distribution:
\begin{equation}
    x_{t} \sim p(x_t | x_1, C_2, x_2, C_3, ..., x_{t-1}, C_{t} ).
\label{eq:sample}
\end{equation}
However, different from the simple case that models only two adjacent views~\cite{Infinite_Nature, geogpt}, sequential modeling poses two problems:
(i) Self-attention alone does not ensure that the relationship between every two patches across space and time are properly modeled, given a large number of input patch tokens increase the optimization difficulty;
(ii) More careful designs should be taken into account to ensure a consistent long-term synthesis.
For the first problem, we propose a Camera-Aware Bias in self-attention as a locality constraint (Sec.~\ref{sec:bias}).
For the second problem, we propose several key techniques for both training and inference (Sec.~\ref{sec:architecture} \& Sec.~\ref{sec:details}).

\subsection{Network Architecture}
\label{sec:architecture}
\noindent\textbf{Overview.} 
Direct learning the distribution in Eq.~\ref{eq:prob_1} in an end-to-end manner is difficult because the model needs to capture interactions inside the sequence and guarantee high-quality generation at the same time.  To tackle this problem, we follow previous methods~\cite{pixelsin, vqgan} to adopt a two-stage training. 
For the first stage, we pretrain a VQ-GAN~\cite{vqgan} mapping the images to ``tokens'', consisting an encoder $E$ that encode images to discrete representations, a decoder $D$ that map the representations to high-fidelity outputs, and a codebook $\mathcal{B} = \{ b_i \}_{i=1}^{|\mathcal{B}|}$ of discrete representations $b_i \in \mathbb{R}^{d_b}$. 
After processing the images into ``tokens'', we use a GPT architecture~\cite{gpt} which modifies the Transformer~\cite{tranformer} architecture with a causal self-attention mask to enable autoregressive generation. 
We then introduce each of the modules in our system in detail, as shown in Figure~\ref{fig:pipeline} (a).

\noindent\textbf{Image Encoder $E$.}
For an input sequence of images $\{x_l \}_{l=1}^{L}$, the $l$-th frame $x_l \in \mathbb{R}^{H\times W \times 3}$ can be converted into the latent space by the pretrained VQ-GAN encoder denoted as:
\begin{equation}
    y_{l} = E(x_{l}),
\end{equation}
where $y_{l} \in \mathbb{R}^{hw \times d_b}$ is the latent variable with $h \times w$ tokens. Note that $h$ and $w$ are smaller than $H$ and $W$, making the size tractable.
Then $y_{l}$ can be quantized to get a sequence of integers $z^{I}_l \in \mathbb{R}^{hw}$ which index the learned codebook $\mathcal{B}$:
\begin{equation}
    z^{I}_{l,k} = \mathop{\arg\min}_{j} \| y_{l, k} - b_j \|^2,
\end{equation}
where $y_{l, k}$ and $z^{I}_{l,k}$ are the $k$-th tokens of $y_{l}$ and $z^{I}_l$.

\noindent\textbf{Image Decoder $D$.}
Given the nearest indexes $z^I_l$, we can decode it back to a high-fidelity image using the pretrained VQ-GAN decoder. 
$z^I_l$ is first embedded by the codebook $\mathcal{B}$:
\begin{equation}
    b_l = B[z^I_l],
\end{equation}
where $b_l \in \mathbb{R}^{(hw) \times d_B}$. Then,  $b_l$ can be decoded to reconstruct the original image:
\begin{equation}
    \hat{x} = D(b_l),
\end{equation}
where $\hat{x} \in \mathbb{R}^{H \times W \times 3}$. In this way, we can model the Eq.~\ref{eq:prob_1} with the discrete representation of images $z^I$ in the latent space of the VQ-GAN. Moreover, the discrete representation is well-aligned with the ``word'' in NLP and thus suitable for efficiently training GPT-like architecture~\cite{vqgan}.

\noindent\textbf{Camera Encoder $E^C$.}
For the camera model, we follow previous work~\cite{synsin,geogpt} to assume it as a pinhole one, such that a desired geometric transformation between two images can be determined by the intrinsic camera matrix $K$, a rotation matrix $R$, and a translation matrix $t$. 

\noindent\textit{Canonical Modeling.}
In our method, to improve consistency, we propose to use \textit{canonical modeling}, such that the first image is assumed as a canonical view. Thus, the input sequence of camera transformation $\{C_l \}_{l=2}^{L}$ is relative to the canonical view, i.e., $C_l = (K, R_{1 \rightarrow l},t_{1 \rightarrow l})$. We encode $C_l$ to latent representation $C^e_l \in \mathbb{R}^{M \times d_e}$ by:
\begin{equation}
    C^e_l = E^C(C_l),
\end{equation}
where the camera parameters inside $C_l$ are flattened and concatenated to shape $M \times 1$ and $E^C$ is a linear layer mapping from $R$ to $R^{d_e}$.

\noindent\textbf{Transformer $\mathcal{T}$.}
Given the encoded images embeddings $\{ z_l \}_{l=1}^L$ and camera embeddings $\{ C^e_l \}_{l=2}^L$, we use a transformer to model the conditional probability in Eq.~\ref{eq:prob_1} in the latent space. 

\noindent\textit{Decoupled Positional Embedding (P.E.).} 
To deal with the spatial-temporal relationship, we propose a \emph{decoupled positional embedding}.
The tokens of the image are first calculated with the consideration of spatial information:
\begin{equation}
    g^I_{l} = \lambda(z_l) + P^I,
\end{equation}
where $\lambda(.)$ is a embedding function that maps $z_l$ into the latent space $\in \mathbb{R}^{hw \times d_e}$ of transformer and $P^I \in \mathbb{R}^{hw \times d_e}$ is the learnable spatial positional embedding sharing across the images.
Similarly, the camera tokens are calculated as:
\begin{equation}
    g^C_{l} = C^e_l + P^C,
\end{equation}
where $P^C \in \mathbb{R}^{M \times d_e}$ is the learnable camera positional embedding sharing across the cameras.
Then, the input tokens to transformer are calculated as:
\begin{equation}
  v = [g^I_{1}, g^C_{2}, ..., g^I_{L-1}, g^C_{L}, g^I_{L}] + P^T,
\end{equation}
where $v \in \mathbb{R}^{N \times d_e}$ is the input tokens to transformer and $P^T \in \mathbb{R}^{N \times d_e}$ is the sinusoidal position embedding indicating order of tokens, modeling temporal relationship.

Now, the transformer $\mathcal{T}$ can be trained in an autoregressive way, denoted as:
\begin{equation}
    h_n = \mathcal{T}(v_{<n}),
\end{equation}
where $h_n$ is the $n$ step of output hidden states $h \in \mathbb{R}^{N\times d_e}$.
In fact, we only select hidden states for image prediction tokens and feed it to a linear layer to get the probability $p^I$:
\begin{equation}
 p^{I}_{l,k} = p(\hat{z}^{I}_{l,k} | v_{<n}) = {\rm softmax}({\rm Linear}(h_n)),
\end{equation}
where $n$ is the index corresponds to the $k$-th token of the $l$-th frame, i.e., $n = (l-1)(hw+M) + k$. And the linear layer maps $\mathbb{R}^{d_e}$ to $\mathbb{R}^{|B|}$.
Finally, the transformer together with the camera encoder are trained using cross-entropy loss, leading to the training objective:
\begin{equation}
\mathcal{L} = \sum_{l = 2}^{L} \sum_{k = 1}^{hw} CE(p^I_{l,k}, z^I_{l,k}),
\end{equation}
where $CE(.)$ calculates the cross-entropy between the probabilities and given labels, and $z^I$ is the corresponding indexes in the codebook $\mathcal{B}$.

\subsection{Camera-Aware Bias in Transformer}
\label{sec:bias}
Self-attention in Transformer captures global dependency, which is a desirable property for novel view synthesis. However, since only self-attention and MLP are applied in the Transformer, there is a lack of inductive bias on 3D~\cite{missing}. When facing thousands of tokens, including information interaction between tokens across spatial and time, it is hard to capture the significant dependencies (e.g., whether two patches should be perceptually consistent) without any constraints and inductive bias. 

An intuitive way to introduce 3D-aware inductive bias is to inject 3D convolutions. In ConvNets, 3D convolution serves as a 3D-aware inductive bias with the constraint on locality in both spatial and time~\cite{I3D}. Thus, it may be beneficial to inject 3D convolutions into Transformers to introduce 3D-aware inductive bias. However, the motion between two adjacent views can be so large that the overlapping pixels in geometric transformation are not in the local window, which cannot be modeled by one time of convolution operation. 
Our key insight to solve this problem is that there is a clear relationship between frames in the video, such that the correspondence between frame $x_i$ and frame $x_j$ is determined by relative camera transformation $(K, R_{i \rightarrow j}, t_{i \rightarrow j})$. 
We can incorporate such spatial-temporal dependency between pixels as a 3D-aware inductive bias in Transformer.
Inspired by the exploration on relative position bias in computing affinity matrix in self-attention based on image coordinate~\cite{swin}, we model the observed relationship as a novel Camera-Aware Bias in self-attention block, as shown in Figure~\ref{fig:pipeline} (b).

Given the query $q_i \in \mathbb{R}^{hw \times d_e}$ corresponding to $i$-th frame, the key and value $k_j, v_j \in \mathbb{R}^{hw \times d_e}$ corresponding to $j$-th frame, we can first compute the similarity matrix $a_{i, j}  \in \mathbb{R}^{hw \times hw}$ with Camera-Aware Bias, denoted as:
\begin{equation}
 a_{i,j} =  q_i k_j + \phi([K, R_{i \rightarrow j}, t_{i \rightarrow j}]),
\end{equation}
where $\phi: \mathbb{R}^{M} \rightarrow \mathbb{R}^{hw \times hw}$ is an MLP.
Then, the self-attention between the hidden states corresponding to $i$-th frame and $j$-th frame is calculated by:
\begin{equation}
{\rm Attention}(q_i, k_j, v_j) = {\rm softmax}(\frac{a_{i,j}}{\sqrt{d_e}})v_{j},
\end{equation}
where ${\rm softmax(.)}$ here also takes similarity for cameras and other frames into account. For the similarity between frames and cameras, we do not apply any bias.
Note that our design is applicable to causal self-attention by setting $j < i$.
By adding the Camera-Aware Bias, each patch will have a stronger bias depending on relevant patches connected by the camera, which serves as a 3D-aware inductive bias.

\subsection{Training and Inference Details}
\label{sec:details}
We then introduce several key techniques for training and inference in our method.

\noindent\textbf{Overlapping Iterative Modeling.}
In our task, we target generating long-term 3D scene video with unconstrained length $T$. However, it is never possible to set the length of the training sequence $L$ to infinity. Thus, we choose an iterative modeling strategy. 
Given a single image $x_1$, we first generate $x_2, ... x_L$ in an autoregressive manner. Then, instead of only using $x_L$, we aggregate information from $x_2, ... x_L$ to generate $x_{L+1}$ and so on. This overlapping iterative modeling allows us to inference for unconstrained length and maintains perceptual consistency.
As we show in Sec.~\ref{sec:ablation}, this strategy is sufficient for a consistent long-term 3D scene video even with a small $L$. 

\noindent\textbf{Error Accumulation.}
As pointed in \cite{RossGB11,Infinite_Nature}, a key challenge in generating long sequences is dealing with the accumulation of errors.
Even a tiny perturbation in each iteration can eventually lead to predictions outside the distribution and thus undesirable results.
For an autoregressive Transformer, though we still need teacher forcing in training, we can partially simulate the error accumulation process during inference. We can first sample the predicted novel views from the predicted logits with the image decoder $D$ and then finetune the model with its own predicted outputs, which improves the visual quality for long-term synthesis as shown in Sec.~\ref{sec:ablation}. 

\noindent\textbf{Beam Search.} 
During inference, we need to sample next frame $x_t$ from Eq.~\ref{eq:sample}. Considering the consistency, we need to choose $x_t$ with the most likely sequences of tokens. 
However, decoding the most likely output sequence is exponential in the length of the output sequence, and thus it is intractable~\cite{Rossi13}.
We find that greedily take the most likely next step as the sequence leads to unnatural artifacts. 
Thus, we adopt a beam search strategy~\cite{RN2020}. 
Starting with the $k$ most likely codes in the VQ codebook as the first step in the sequence, we expand the top $k$ possible next steps instead of all possible in original algorithms for faster speed. Then we keep the $k$ most likely ones and repeat. In this way, we find a more optimal sample than a greedy search.

\section{Experiments}
In this section, we provide an empirical evaluation of our method. We demonstrate the power of our approach with an autoregressive Transformer on the view synthesis task.

\subsection{Experimental setup}

\noindent\textbf{Datasets.} 
We follow the common protocol~\cite{synsin,pixelsin,video_autoencoder} to evaluate our method on \textit{Matterport3D}~\cite{MP3D} and \textit{RealEstate10K}~\cite{ZhouTFFS18}. Matterport3D consists of 3D models of scanned and reconstructed building-scale scenes, of which $61$ are for training and $18$ are for testing. To generate long-term episodes, we use an embodied agent in Habitat~\cite{habitat19iccv} from one point in the scene to another point. In total, we render $6000$ videos for training and $500$ videos for testing.
RealEstate10K is a collection of videos of footages of real estates (both indoor and outdoor). We follow \cite{video_autoencoder} to use $10,000$ videos for training and $5,000$ videos for testing.

\begin{table}[t]
\begin{center}
\resizebox{\linewidth}{!}{
\begin{tabular}{lcccc}
\toprule
\multicolumn{1}{l}{Method} &  \multicolumn{2}{c}{Matterport3D} & \multicolumn{2}{c}{RealEstate10K} \\
 & LPIPS $\downarrow$  & PSNR $\uparrow$  & LPIPS $\downarrow$  & PSNR $\uparrow$ \\
\midrule
SynSin~\cite{synsin} & 3.53 & 13.92 & 2.55 & 14.77\\
SynSin-6x~\cite{pixelsin}  & 3.59 & 14.33 & 2.62 &  14.89\\
GeoGPT~\cite{geogpt} & 3.09 & 15.24 & 2.68 & 14.42\\
Ours & \textbf{2.97} & \textbf{16.06} & \textbf{2.53} & \textbf{15.60} \\
\bottomrule
\end{tabular}
}
\end{center}
\vspace{-1em}
\caption{Quantitative evaluation on \textit{short-term} view synthesis.}
\label{tbl:short}
\vspace{-1em}
\end{table}

\noindent\textbf{Baselines.} 
We compare our approach with three state-of-the-art single-image novel view synthesis work: \textit{SynSin}~\cite{synsin}, \textit{PixelSynth}
\footnote{The current implementation of PixelSynth only supports $10$ discrete directions.  We compare against it in Sec.~\ref{sec:long} following their setting.}~\cite{pixelsin} and \textit{GeoGPT}~\cite{geogpt}. 
SynSin and PixelSynth utilize point cloud as a geometric representation.
We also adopt an improved version of SynSin, named \textit{SynSin-6x}, provided by \cite{pixelsin}, which trained on larger view change.
GeoGPT is a geometry-free method with probabilistic modeling between two adjacent views. We provide comparisons with additional baselines including~\cite{Infinite_Nature,video_autoencoder} in appendix Sec.~\ref{sec:supp_baseline}. Specifically, for comparing to the Infinite Nature method proposed by Liu et al.~\cite{Infinite_Nature}, as we are focusing on different settings and the code of ~\cite{Infinite_Nature} is not publicly available, we compare to an alternative approximation as suggested by~\cite{pixelsin}. 

\noindent\textbf{Implementation Details.}
For preprocessing, we resize all images into a resolution of $H \times W = 256 \times 256$.
For our experiments on both Matterport3D and RealEstate10K, we adopted the VQ-GAN from \cite{geogpt} pretrained on RealEstate10K. The number of entries in the codebook $\mathcal{B}$ is $16384$.
For the Transformer, we adopt a GPT-like architecture~\cite{gpt} with a stack of $32$ transformer blocks containing casual self-attention modules. 
During training, the training video clip consists $L = 3$ frames, which will be discussed in Sec.~\ref{sec:ablation}. The encoded image is of shape $h \times w = 16 \times 16$ and the camera embedding is of length $M = 30$, which lead the total sequence length $N = 828$. 
We train our Transformer using a batch size of $16$ for $200K$ iterations with an AdamW optimizer~\cite{AdamW} (with $\beta_1 = 0.9$, $\beta_2 = 0.95$). We set the initial learning rate to $1.5 \times 10^{-4}$ and apply a cosine-decay learning rate schedule~\cite{decay} towards zero. 
For beam search, we set $k = 3$.
We defer more details to the supplementary material.

\subsection{Evaluation on Short-Term View Synthesis}
We evaluate our method against the baselines on short-term view synthesis in the considered range of previous novel view synthesis methods. 
In this setting, we adopt the standard metrics in view synthesis task: PSNR and LPIPS~\cite{zhang2018perceptual}.
PSNR measures pixel-wise differences between two images, and LPIPS measures the perceptual similarity in deep feature space.
As pointed by \cite{pixelsin}, PSNR and LPIPS also measure \emph{consistency} for a unimodal task, such as the short-term view synthesis.
For both datasets, we randomly select test sequences with an input frame and $5$ subsequent ground-truth frames.

Table~\ref{tbl:short} shows the quantitative results for our method. Without an intermediate geometry, our method can still outperform the methods with explicit geometric modeling in terms of short-term view synthesis.
Moreover, our method also outperforms the geometry-free baseline, GeoGPT, by a large margin since this method does not ensure consistency.

\subsection{Evaluation on Long-Term View Synthes}
\label{sec:long}

\begin{table}[t]
\begin{center}
\resizebox{\linewidth}{!}{
\begin{tabular}{lcccc}
\toprule
\multicolumn{1}{l}{Method} &  \multicolumn{2}{c}{Matterport3D} & \multicolumn{2}{c}{RealEstate10K} \\
 & A/B vs. Ours  & FID$\downarrow$  & A/B vs. Ours & FID$\downarrow$ \\
\midrule
SynSin~\cite{synsin} & 82.0\% & 152.51 & 92.5 \% & 75.47\\
SynSin-6x~\cite{pixelsin} & 87.0\% & 153.96 & 88.5 \% & 48.71\\
GeoGPT~\cite{geogpt} & 81.5 \% & 99.06 & 68.5 \% & 53.82 \\
Ours & -- &  \textbf{57.22} & -- &
\textbf{32.88} \\
\midrule 
PixelSynth~\cite{pixelsin} & 69.0\% & 146.54 &  63.0\% & 98.87\\
Ours*  & -- &  \textbf{75.96} & -- & \textbf{82.51} \\
\bottomrule
\end{tabular}}
\end{center}
\vspace{-1.2em}
\caption{Image quality and scene consistency evaluation on \textit{long-term} view synthesis.
For the A/B test, each cell lists the fraction of pairwise comparisons in which scenes synthesized by our approach were rated more consistent than scenes synthesized by the corresponding baseline. 
Ours* indicates our results following the setting of PixelSynth.
}
\label{tbl:long}
\vspace{-0.5em}
\end{table}

\begin{table}[t]
\begin{center}
\resizebox{\linewidth}{!}{
\begin{tabular}{lcccc}
\toprule
\multicolumn{1}{l}{Method} &  \multicolumn{2}{c}{Matterport3D} & \multicolumn{2}{c}{RealEstate10K} \\
 & LPIPS $\downarrow$  & PSNR $\uparrow$  & LPIPS $\downarrow$  & PSNR $\uparrow$ \\
\midrule
SynSin~\cite{synsin} & 3.85 & 13.51 & 3.41 & 12.18\\
SynSin - 6x~\cite{pixelsin} & 3.85 & \textbf{14.03} & 3.42 & 12.28\\
GeoGPT~\cite{geogpt} & 3.71 & 11.43 & 3.44 & 10.61\\
Ours & \textbf{3.54} & 12.89 & \textbf{3.20} & \textbf{12.36} \\
\bottomrule
\end{tabular}}
\end{center}
\vspace{-1.2em}
\caption{Quantitative evaluation on \textit{long-term} view synthesis. 
Though PSNR and LPIPS are poor metrics for extrapolation tasks~\cite{pixelsin,KrishnanTSMLBF19}, we report them for reference.}
\label{tbl:long_tradition}
\vspace{-0.5em}
\end{table}

\begin{figure}[t]
\centering
\includegraphics[width=\linewidth]{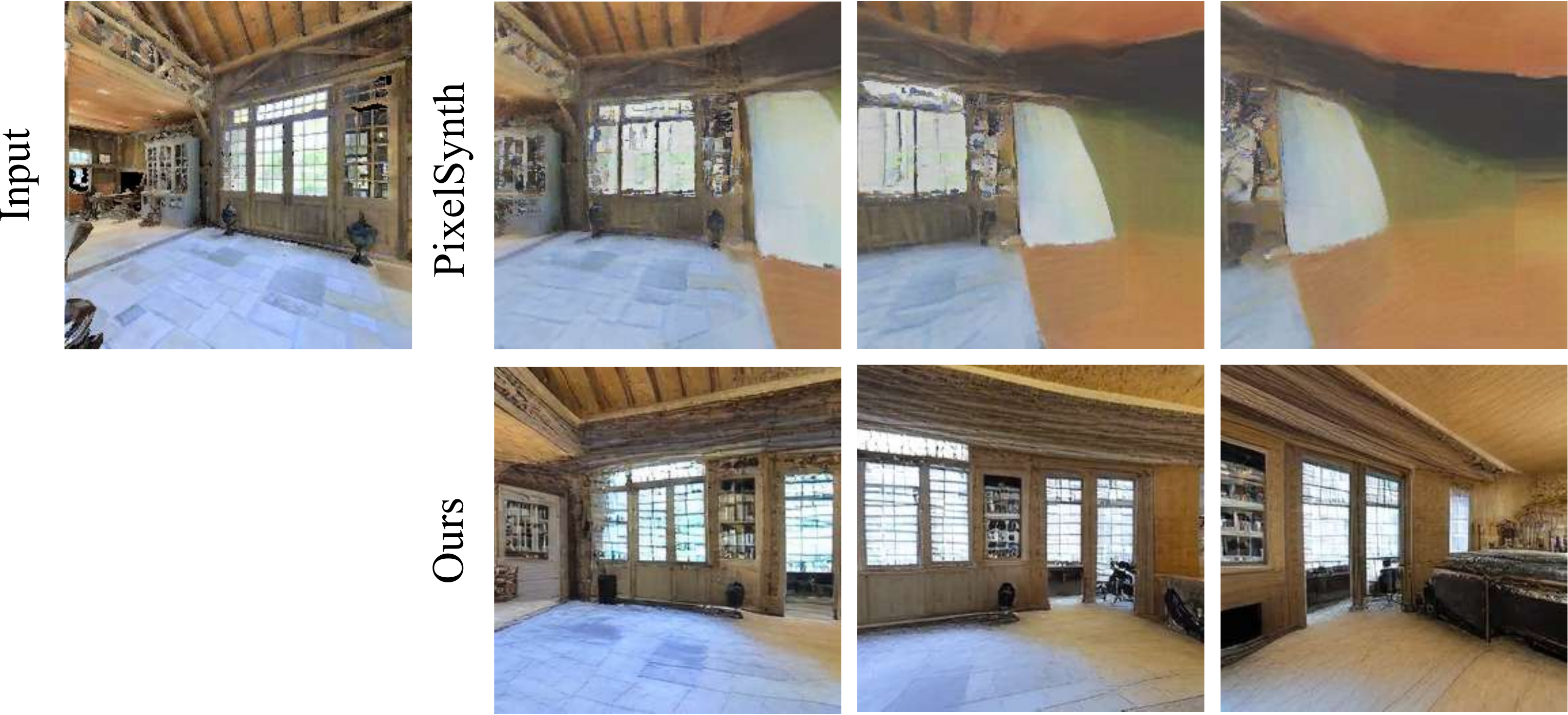}
\vspace{-1.2em}
\caption{Qualitative comparison between our method and PixelSynth~\cite{pixelsin}. Though PixelSynth conducts outpainting explicitly, it is not capable of synthesizing a long-range view. Sequential modeling instead creates realistic and consistent views since it facilitates outpainting conditioned on pixels from previous frames.}
\label{fig:qual_pixel}
\vspace{-1em}
\end{figure}

We then evaluate our method on the \emph{long-term} view synthesis task. 
Prior work~\cite{KrishnanTSMLBF19, pixelsin} points out that PSNR and LPIPS are poor metrics for scene extrapolation tasks cause there are multiple possibilities for the output. 
Thus, for image quality, we follow \cite{Infinite_Nature} to use FID~\cite{HeuselRUNH17}, which is a distribution-level similarity measurement between generated images and real images.
For consistency, we follow \cite{pixelsin} to conduct user study on Amazon Mechanical Turk following the A/B test protocol~\cite{ChenK17}. Each user is presented with a video generated by our method and a baseline simultaneously during the user study. Then the user needs to choose a more consistent one. 
For both datasets, we randomly select test sequences with an input frame and $20$ subsequent GT frames with significant camera motion, of which each covers an extended range of footage.
To compare with PixelSynth, we randomly sample an input frame and several outpainting directions to form a test sequence.

\begin{figure*}[t]
\centering
\includegraphics[width=\linewidth]{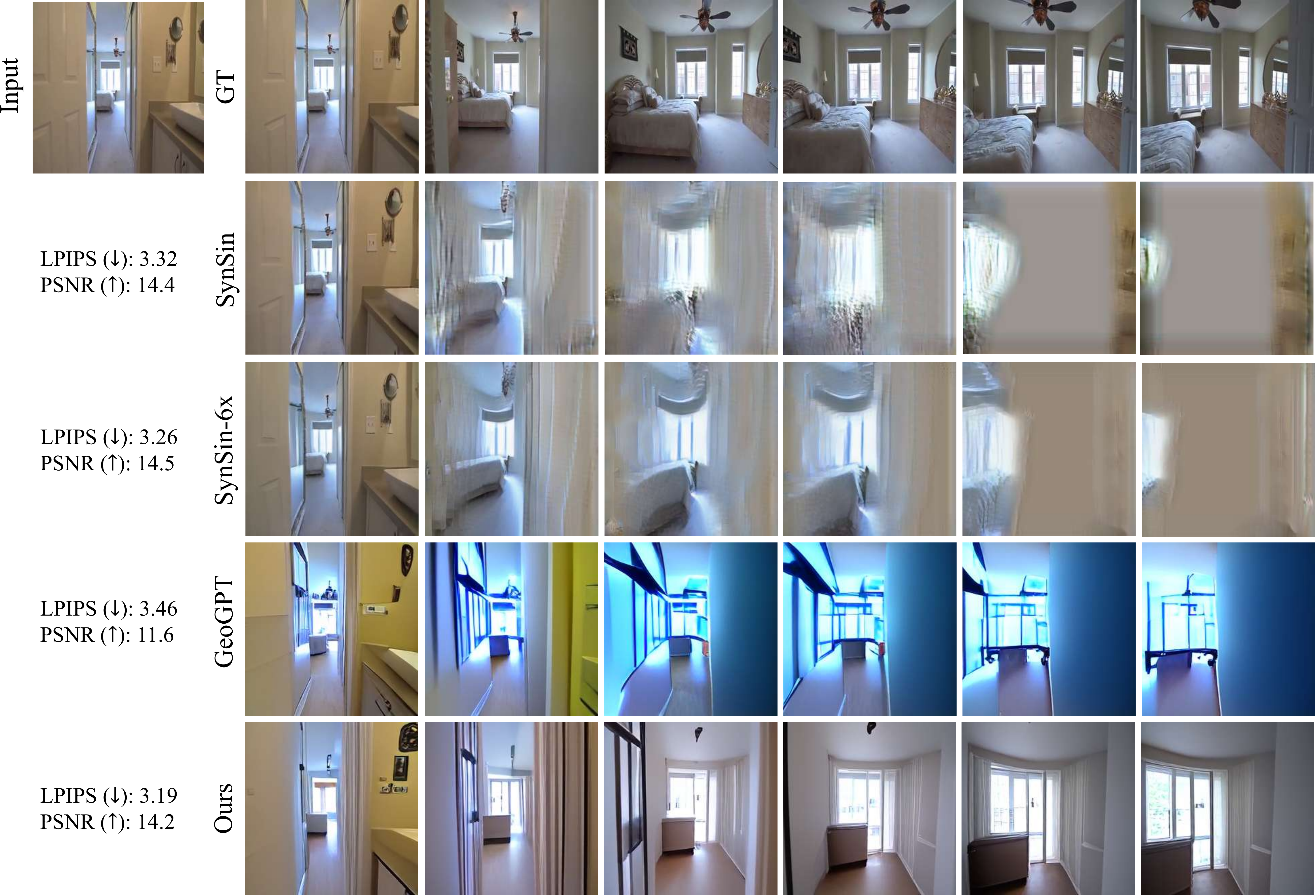}
\vspace{-1em}
\caption{Qualitative comparison between our method and baselines in terms of \textit{long-term} view synthesis. 
Prior work is not capable of synthesizing a consistent long-term scene video. Though our result is not the same as ground truth, it is perceptual consistency and of high-fidelity. For more results, please refer to supplementary materials.}
\label{fig:qual_long}
\end{figure*}

We report the quantitative comparisons in Table~\ref{tbl:long}. 
For both image quality and consistency, our method is significantly better than other baselines, including geometry-based and geometry-free ones. 
This is consistent with qualitative results, as shown in Figure~\ref{fig:qual_pixel} and Figure~\ref{fig:qual_long}.
On Matterport3D, the gap is even more prominent due to the view angle changes being more significant.
Notably, geometry-free methods achieve better image quality on long-term view synthesis. SynSin-6x performance is still not good, indicating that training previous methods on larger camera changes helps but does not account for the main issue.
In addition, we follow past work to report PSNR and LPIPS in Table~\ref{tbl:long_tradition}, which are poor measures for extrapolation tasks~\cite{pixelsin, KrishnanTSMLBF19}. For example, though SynSin-6x usually produces entirely gray results, as shown in Figure~\ref{fig:qual_long}, its PSNR is good.

\begin{figure}[t]
\centering
\begin{tabular}{c@{\hspace{0.3em}}c@{\hspace{0.3em}}c}
{\includegraphics[width=0.32\linewidth]{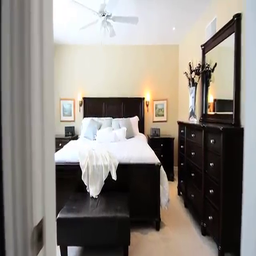}} &
{\includegraphics[width=0.32\linewidth]{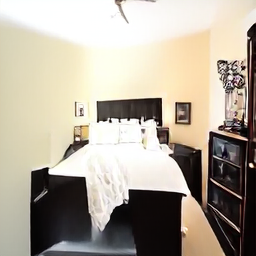}} &
{\includegraphics[width=0.32\linewidth]{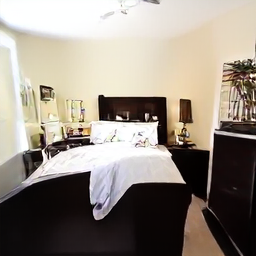}}
\\
Input & Ours & Ours w/o Bias \\
\end{tabular} 
\vspace{-1em}
\caption{Visual ablation study. The proposed camera-aware bias benefits both the consistency between frames and image quality.
}
\label{fig:ablation}
\end{figure}

\subsection{Ablation Study}
\label{sec:ablation}
We report some ablations of our method in terms of long-term view synthesis on the Matterport3D dataset.

\begin{table}[t]
\begin{center}
\resizebox{\linewidth}{!}{
\begin{tabular}{l@{\hspace{3em}}c@{\hspace{3em}}c}
\toprule
\multicolumn{1}{l}{Method} & A/B vs. Ours & FID$\downarrow$ \\
\midrule
Ours (Full Model)& -- & \textbf{57.22} \\
\qquad -- Decoupled P.E. & 65.0\% &  70.47 \\
\qquad  -- Camera-Aware Bias & 73.8\% & 60.42\\
\qquad  -- Error Accumulation & 56.3\% & 66.81 \\
\bottomrule
\end{tabular}
}
\end{center}
\vspace{-1.2em}
\caption{Ablation study on Matterport3D in terms of \textit{long-term} view synthesis. We ablate aspects of our model to investigate their influence on the results.}
\label{tbl:abl}
\vspace{-1em}
\end{table}

\noindent\textbf{Camera-Aware Bias.} 
As shown in Figure~\ref{fig:ablation}, the Camera-Aware Bias improves the image quality and the consistency between frames. 
Table~\ref{tbl:abl} also confirms this observation, indicating that bringing locality into autoregressive Transformer is critical, especially for consistency. 

\noindent\textbf{Decoupled positional embedding.}
We replace our decoupled positional embedding (P.E.) with a vanilla
learnable positional embedding. As shown in Table~\ref{tbl:abl}, both image quality and consistency drop. 

\noindent\textbf{Error accumulation.}
As shown in Table~\ref{tbl:abl}, finetuning the model by stimulating error accumulation benefits the long-term view synthesis.

\noindent\textbf{Length of video clips.} 
We compare our default length of video clips with variants that modify the length during training. As shown in Table~\ref{tbl:clip}, the consistency improves significantly when the length increase from $2$ to $3$.  When the length further increases to $5$, the consistency remains nearly unchanged. For the image quality, there is a significant drop when we expand the length to $5$. We hypothesize that the numbers of tokens are too large that the Transformer is difficult to optimize. Considering the computation resource and performance, we set the length of video clips to $3$.

\begin{table}[t]
\begin{center}
\resizebox{0.9\linewidth}{!}{
\begin{tabular}{l@{\hspace{2em}}c@{\hspace{2em}}c@{\hspace{2em}}c@{\hspace{2em}}c}
\toprule
 & L=2 & L=3 & L=4 & L=5\\
\midrule
FID$\downarrow$ & 70.62 & \textbf{57.22} &  62.34 &  93.85 \\
A/B vs. Ours & 97.0\% & -- & 54.0\% & 49.0\% \\ 
\bottomrule
\end{tabular}
}
\end{center}
\vspace{-1.5em}
\caption{Ablation study on length of video clips $L$.}
\label{tbl:clip}
\vspace{-1em}
\end{table}

\section{Discussion}
\noindent\textbf{Conclusion.}
We propose an autoregressive Transformer based model to solve novel view synthesis, especially when synthesizing long-term future in indoor 3D scenes. This method leverages a locality constraint based on the input cameras in self-attention to ensure consistency among generated frames. Our method can get superior performance in novel view synthesis compared to the state-of-the-art approaches. 
To conclude, we take a further step to explore the capabilities of geometry-free methods and manage to synthesize consistent high-fidelity 3D scenes.

\noindent\textbf{Limitations and Future Work.}
Nevertheless, there are challenges remain. 
First, the current inference speed of the autoregressive models is slightly slower than vanilla models (details in appendix Sec.~\ref{sec:time}). Further advancements in the autoregressive model still call for need.
Second, current metrics like PSNR and LPIPS are not perfect to evaluate long-term view synthesis. New metrics for this task deserve more attention.

{\small
\bibliographystyle{ieee_fullname}
\bibliography{egbib}
}

\clearpage
\appendix

\section{Additional View Synthesis Results}

\subsection{Our Qualitative Results}

\begin{figure*}[h]
\centering
\begin{tabular}{c@{\hspace{1em}}c}
{\includegraphics[height=0.125\linewidth]{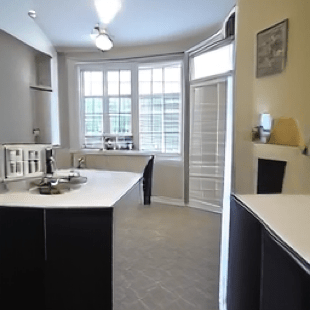}} &
{\includegraphics[height=0.125\linewidth]{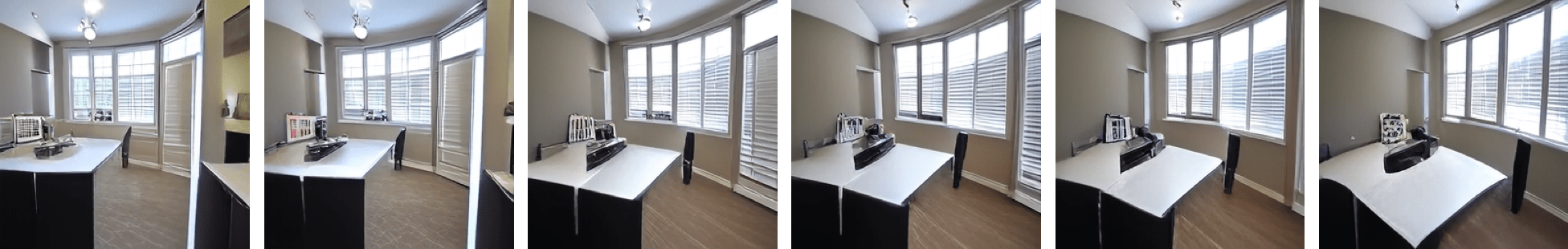}} 
\\
{\includegraphics[height=0.125\linewidth]{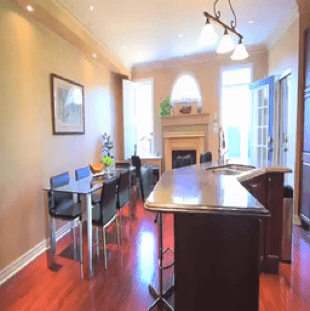}} &
{\includegraphics[height=0.125\linewidth]{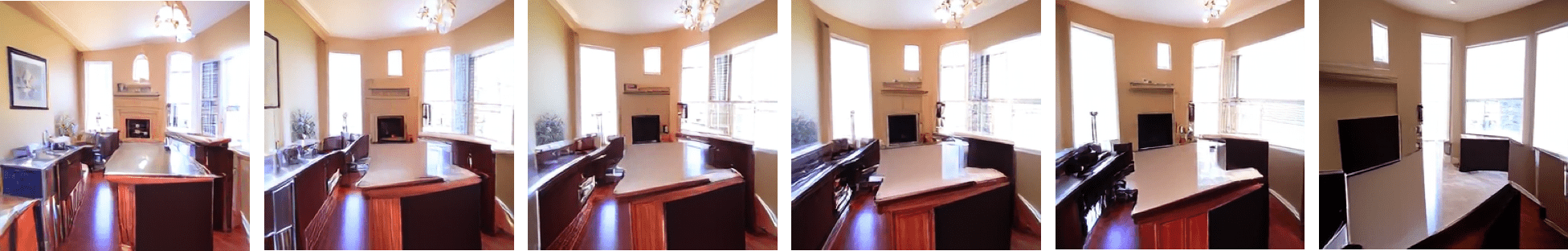}} 
\\
{\includegraphics[height=0.125\linewidth]{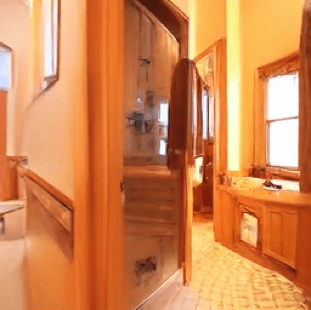}} &
{\includegraphics[height=0.125\linewidth]{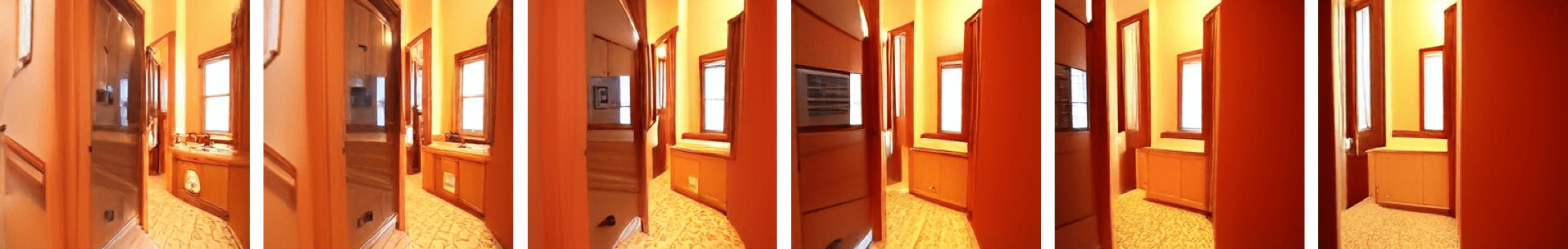}} 
\\
{\includegraphics[height=0.125\linewidth]{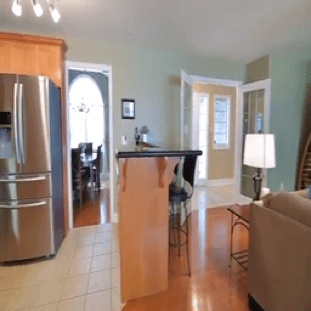}} &
{\includegraphics[height=0.125\linewidth]{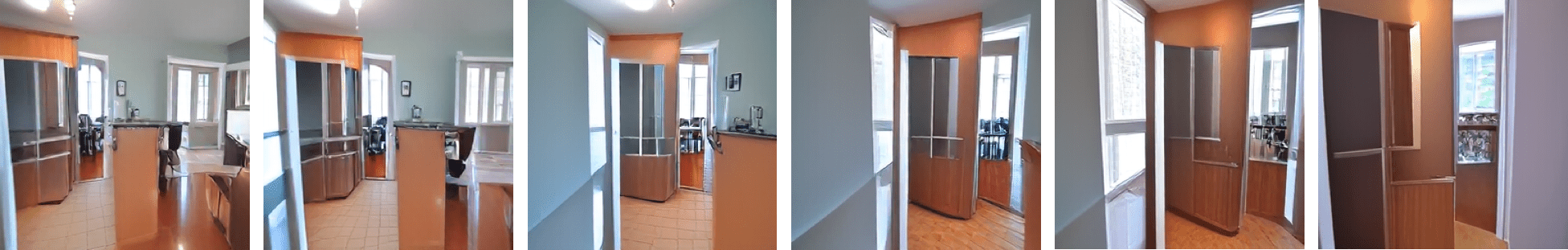}} 
\\
Input & Our Prediction \\
\end{tabular} 
\caption{\textbf{Long-term view synthesis} on RealEstate10K~\cite{ZhouTFFS18}.  Our method is able to synthesize consistent novel views with large camera transformations while maintaining high fidelity.
}
\label{fig:realestate}
\end{figure*}

\begin{figure*}[h]
\centering
\begin{tabular}{c@{\hspace{1em}}c}
{\includegraphics[height=0.125\linewidth]{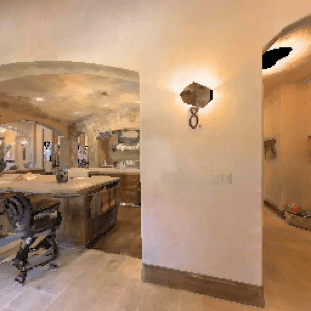}} &
{\includegraphics[height=0.125\linewidth]{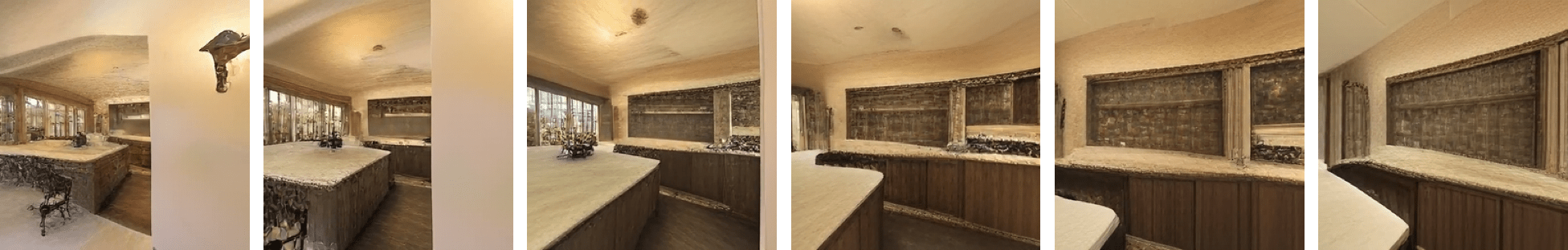}} 
\\
{\includegraphics[height=0.125\linewidth]{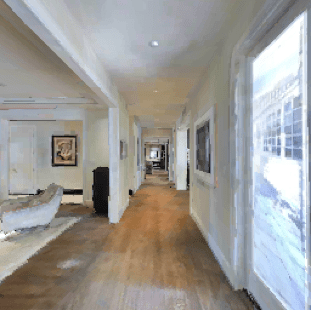}} &
{\includegraphics[height=0.125\linewidth]{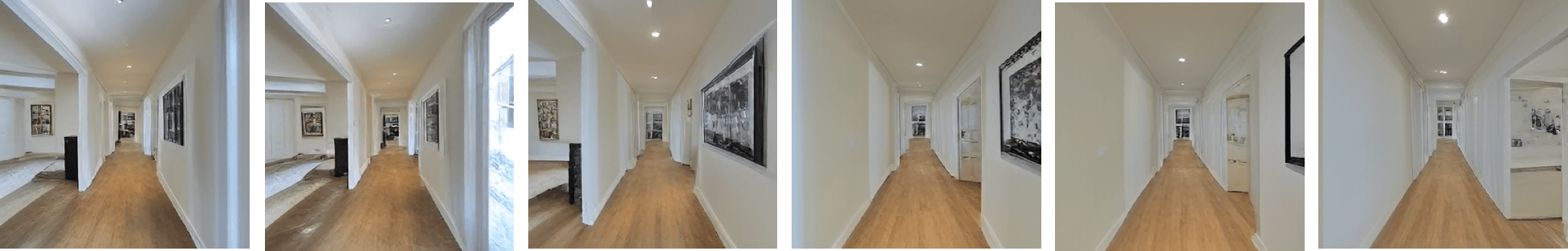}} 
\\
{\includegraphics[height=0.125\linewidth]{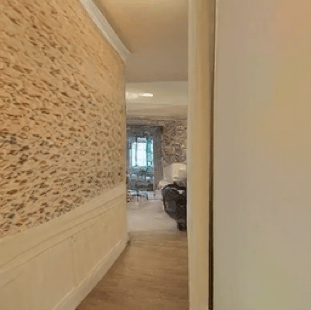}} &
{\includegraphics[height=0.125\linewidth]{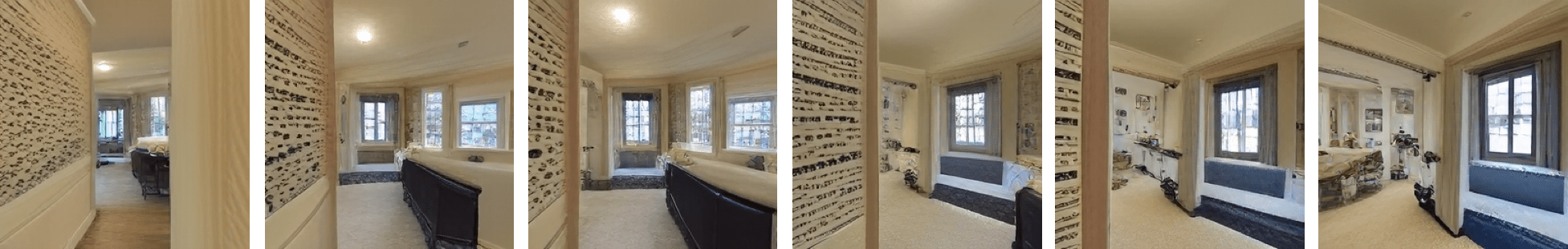}} 
\\
{\includegraphics[height=0.125\linewidth]{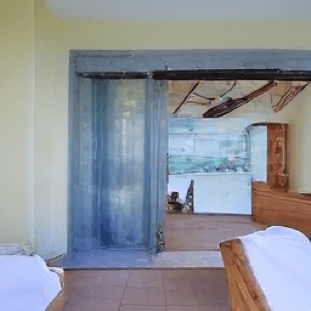}} &
{\includegraphics[height=0.125\linewidth]{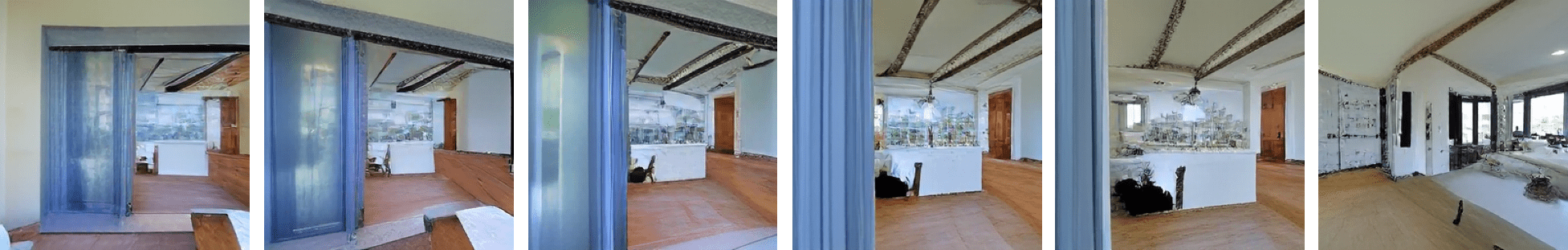}} 
\\
Input & Our Prediction \\
\end{tabular} 
\caption{\textbf{Long-term view synthesis} on Matterport3D~\cite{MP3D}.  Our method is able to synthesize consistent novel views with large camera transformations while maintaining high fidelity.}
\label{fig:mp3d}
\end{figure*}

Figure~\ref{fig:realestate} and \ref{fig:mp3d} provide additional long-term 3D scene videos synthesized by our methods. Our method is able to synthesize consistent novel views with large camera transformations while maintaining high fidelity.

\subsection{Qualitative Comparison with Baselines}
\begin{figure*}[h]
\centering
\begin{tabular}{c@{\hspace{1em}}c}
{\includegraphics[height=0.125\linewidth]{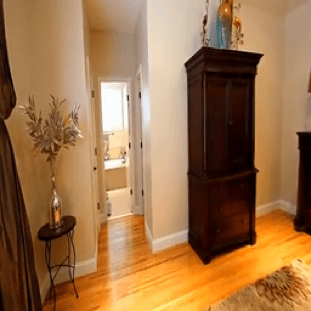}} &
{\includegraphics[height=0.125\linewidth]{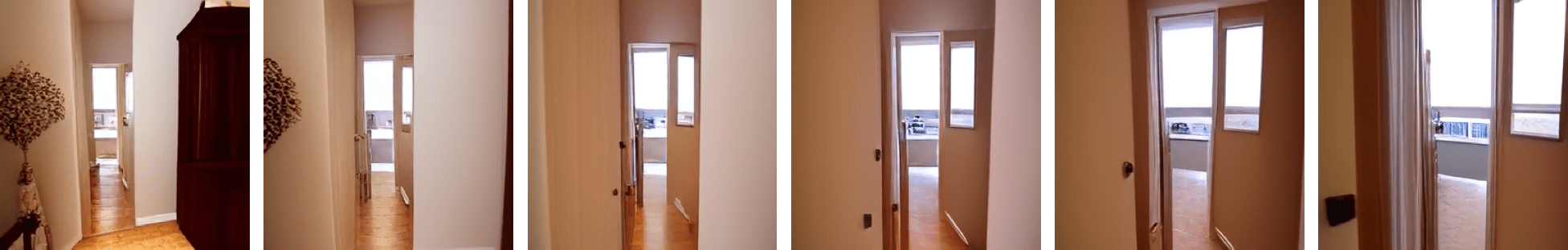}} 
\\
Input & Ours \\
&
{\includegraphics[height=0.125\linewidth]{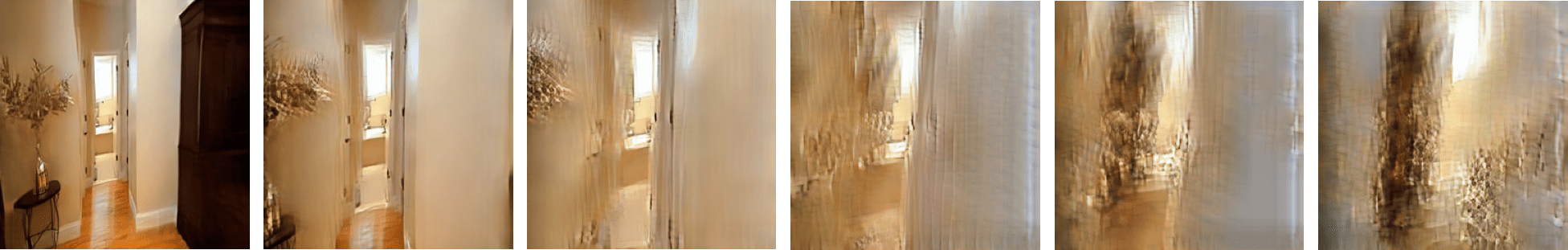}} 
\\
 & SynSin~\cite{synsin} \\
&
{\includegraphics[height=0.125\linewidth]{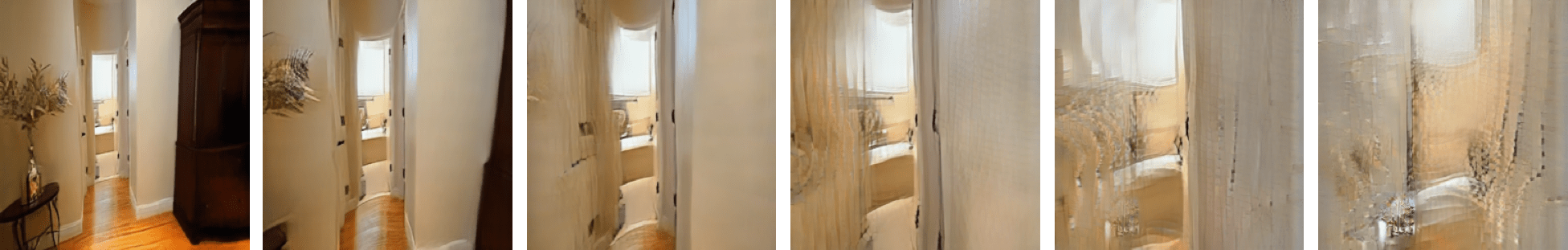}} 
\\
 & SynSin-6x~\cite{pixelsin} \\
&
{\includegraphics[height=0.125\linewidth]{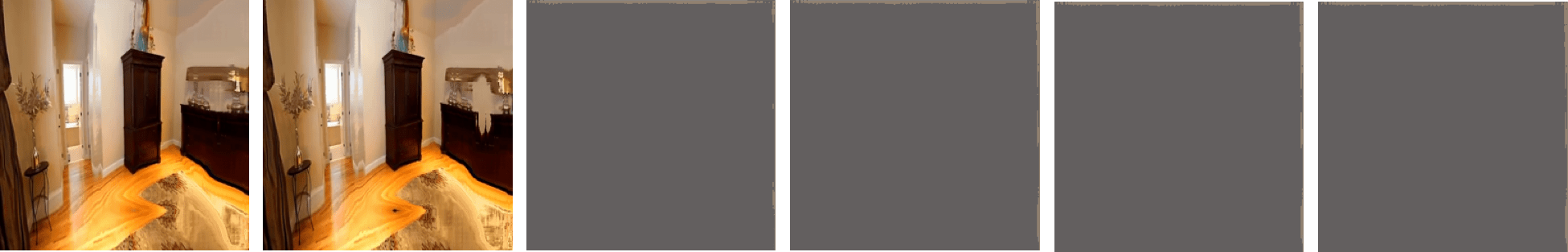}}  
\\
 &  Appearance Flow~\cite{ZhouTSME16} \\
 &
{\includegraphics[height=0.125\linewidth]{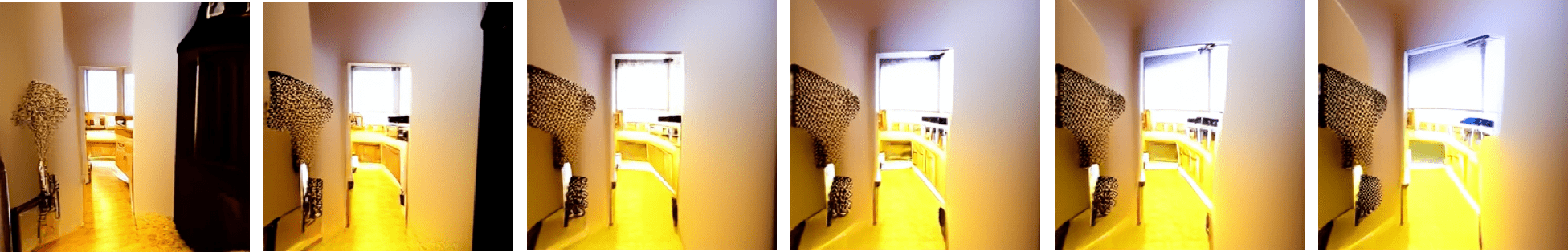}} 
\\
 &  GeoGPT~\cite{geogpt}\\
\end{tabular} 
\caption{\textbf{Long-term view synthesis compared with baselines.}
Previous methods are not capable of synthesizing a consistent long-term scene video. Our method can synthesize long-term views of perceptual consistency and high-fidelity.}
\label{fig:comp_real}
\end{figure*}

Figure~\ref{fig:comp_real}  provide additional comparison with previous methods, including SynSin~\cite{synsin}, SynSin-6x~\cite{pixelsin}, GeoGPT~\cite{geogpt} and Appearance Flow~\cite{ZhouTSME16}. 
The details of the baselines are introduced in Sec.~\ref{sec:baseline}. Our method is able to generate more consistent and clear. 

\subsection{Additional Visual Ablation Study}

Figure~\ref{fig:ablation} provides additional visual ablation study to validate the effectiveness of beam search strategy.

\begin{figure}[h]
\centering
\begin{tabular}{c@{\hspace{0.3em}}c@{\hspace{0.3em}}c}
{\includegraphics[width=0.32\linewidth]{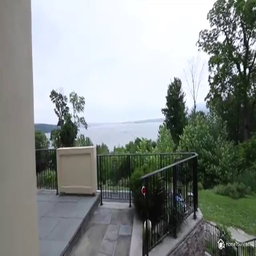}} &
{\includegraphics[width=0.32\linewidth]{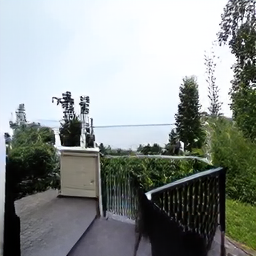}} &
{\includegraphics[width=0.32\linewidth]{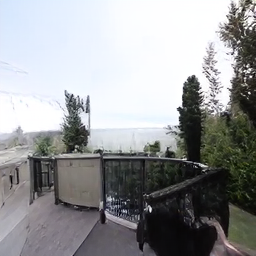}}
\\
Input & Ours & Ours w/o Beam \\
\end{tabular} 
\caption{Visual ablation study on Beam Search.
}
\vspace{-0.2in}
\label{fig:ablation}
\end{figure}
\vspace{-0.1in}

\section{Additional Experiment}

\subsection{Comparison with Additional Baselines}
\label{sec:supp_baseline}

\textbf{Infinite Nature}~\cite{Infinite_Nature}. Our paper focuses on indoor scenes while Infinite Nature proposed by Liu et al.~\cite{Infinite_Nature} focuses on nature scenes and the training code is currently not available online. Our problem is also more challenging given more structural constraints in indoor scenes. As an approximation,  following the suggestion by Rockwell et al.~\cite{pixelsin}, we compare to a method applying SynSin~\cite{synsin} in a sequential manner, namely \textbf{SynSin-Sequential}.
We report the FID results on Matterport3D in Table~\ref{tbl:appendix_comp}. We achieve significant improvements on image quality. 

\textbf{Video Antoencoder}~\cite{video_autoencoder}.
We also compare to Video Autoencoder proposed by Lai et al.~\cite{video_autoencoder} on Matterport3D. As shown in Table~\ref{tbl:appendix_comp}, it performs worse than our method. However, it is worthy to note that Video Autoencoder does not require camera ground-truths during training, which is a more challenging setting.

\begin{table}[h]
\begin{center}
\resizebox{\linewidth}{!}{
\begin{tabular}{lccc}
\toprule
 & Video Antoencoder & SynSin-Sequential & Ours\\
\midrule
FID$\downarrow$ & 229.68 & 158.31 & \textbf{57.22} \\
\bottomrule
\end{tabular}
}
\end{center}
\vspace{-1.5em}
\caption{Comparison on Matterport3D.}
\label{tbl:appendix_comp}
\vspace{-1.5em}
\end{table}

\subsection{Time consumption.}
\label{sec:time}

We measure the average time to generate a frame during inference, as shown in Table~\ref{tbl:time}. 

\begin{table}[h]
\begin{center}
\vspace{-0.5em}
\resizebox{0.85\linewidth}{!}{
\begin{tabular}{l@{\hspace{3em}}c@{\hspace{3em}}c@{\hspace{3em}}c}
\toprule
 & PixelSynth & GeoGPT & Ours\\
\midrule
Time (sec/image) & 24.71 & 8.73 & 13.13 \\
\bottomrule
\end{tabular}
}
\end{center}
\vspace{-1.5em}
\caption{\small{Average inference time (sec/image).}}
\vspace{-1.5em}
\label{tbl:time}
\end{table}

\section{More Implementation Details}

We provide more implementation details of our method.

\textbf{Transformer.}
We follow GPT-2 architecture~\cite{radford2019language} to implement our Transformer. We set the hidden dimension $d_e$ to $1024$, set the number of attention heads to $16$, and use a two-layer MLP with hidden size of $4096$ inside each transformer block. 
For an autoregressive Transformer, we adopt the teacher-force strategy~\cite{gpt} with autoregressive masks during training to enable parallel computing.

\textbf{VQ-GAN.}
We adopt the architecture and training strategy from \cite{vqgan}~\footnote{\url{https://github.com/CompVis/taming-transformers}} for our VQ-GAN part. And we use a downsampling factor of $16$, such that an image of resolution $256 \times 256$ is encoded to $16 \times 16$ tokens.

\section{Details of Baselines}
\label{sec:baseline}

\noindent\textbf{SynSin}\cite{synsin}.
SynSin utilizes a point cloud as an intermediate geometric representation. 
We also consider a baseline, \textbf{SynSin-6x}, which is a version of SynSin trained on much larger view changes. However, these two baselines can only perform inpainting and can not generalize to large view changes.
We adopt the official implementation\footnote{\url{https://github.com/facebookresearch/synsin}}.

\noindent\textbf{PixelSynth}~\cite{pixelsin}.
Based on SynSin, PixelSynth proposes to perform outpainting with the help of VQ-VAE2 and auto-regressive model~\cite{PixelCNN}. 
However, though it can perform outpainting, it still can not apply to the long-term view synthesis as our method does. 
For the implementation, we adopt the official one\footnote{\url{https://github.com/crockwell/pixelsynth}}.

\noindent\textbf{GeoGPT}~\cite{geogpt}. 
GeoGPT is a geometry-free method, which models two adjacent views as a probabilistic model. 
However, GeoGPT can not ensure consistency and does not explore the locality constraint in the autoregressive Transformer.
For the implementation, we adopt the official one\footnote{\url{https://github.com/CompVis/geometry-free-view-synthesis}}. 

\noindent\textbf{Appearance Flow}\cite{ZhouTSME16}.
Besides the baselines used in the main paper, we also compare our method with Appearance Flow, which is also a geometry-free baseline.
Appearance Flow predicts a flow field that warps the original image into a novel view. However, this method can not work well on large camera changes since there are large missing areas after warping. We adopt the implementation provided by SynSin.

\clearpage

\clearpage

\end{document}